# AI-Driven Solutions for Falcon Disease Classification: Concatenated ConvNeXt cum EfficientNet AI Model Approach

Alavikunhu Panthakkan
College of Engineering and IT
University of Dubai
Dubai, U.A.E
apanthakkan@ud.ac.ae

Zubair Medammal
Department of Zoology
University of Calicut
Kerala, India
zubairm@uoc.ac.in

S M Anzar
Department of Electronics
KM College of Engineering
Kerala, India
anzarsm@tkmce.ac.in

Fatma Taher
NextGen Center Academic Director
Zayed University
Dubai, U.A.E
Fatma.Taher@zu.ac.ae

Hussain Al-Ahmad
Vice President for Academic Affairs
University of Dubai
Dubai, U.A.E
halahmad@ud.ac.ae

*Abstract*— **Falconry, an ancient practice of training and hunting with falcons, emphasizes the need for vigilant health monitoring to ensure the well-being of these highly valued birds, especially during hunting activities. This research paper introduces a cutting-edge approach, which leverages the power of Concatenated ConvNeXt and EfficientNet AI models for falcon disease classification. Focused on distinguishing 'Normal,' 'Liver,' and 'Aspergillosis' cases, the study employs a comprehensive dataset for model training and evaluation, utilizing metrics such as accuracy, precision, recall, and f1-score. Through rigorous experimentation and evaluation, we demonstrate the superior performance of the concatenated AI model compared to traditional methods and standalone architectures. This novel approach contributes to accurate falcon disease classification, laying the groundwork for further advancements in avian veterinary AI applications.**

*Keywords—— Falcon Diseases, Health Monitoring, AI-Driven Solutions, Artificial Intelligence, Disease Classification*

## I. INTRODUCTION

### A. About Falcon

Falcons are birds of prey known for their speed, agility, and keen eyesight, found on every continent except Antarctica. They have a streamlined body, sharp talons, and a hooked beak, and are adapted for hunting, using a technique called "contour-hunting" to catch other birds mid-air. Falcons are some of the fastest birds globally, reaching incredible speeds during their hunting dives. They possess exceptional eyesight, adapted to detect ultraviolet light, enabling them to spot prey from great distances. Falcons are birds of prey in the genus Falco, which includes about 40 species. Adult falcons have thin, tapered wings, which enable them to fly at high speed and change direction rapidly. Fledgling falcons, in their first year of flying, have longer flight feathers, which make their configuration more like that of a general-purpose bird such as a broad wing. This makes flying easier while learning the exceptional skills required being effective hunters as adults.

The falcons are the largest genus in the Falconinae subfamily of Falconidae, which itself also includes another subfamily comprising caracaras and a few other species. All these birds kill with their beaks, using a tomial "tooth" on the side of their beaks—unlike the hawks, eagles, and other birds of prey in the Accipitridae, which use their feet. The largest falcon is the Gyrfalcon at up to 65 cm in length. The smallest falcon species is the Pygmy falcon, which measures just 20 cm. As with hawks and owls, falcons exhibit sexual dimorphism, with the females typically larger than the males, thus allowing a wider range of prey species. Some small falcons with long, narrow wings are called "Hobbies" and some which hover while hunting are called "Kestrels. As is the case with many birds of prey, falcons have exceptional powers of vision; the visual acuity of one species has been measured at 2.6 times that of a normal human. Peregrine falcons have been recorded diving at speeds of 320 km/h, making them the fastest-moving creatures on Earth; the fastest recorded dive attained a vertical speed of 390 km/h.

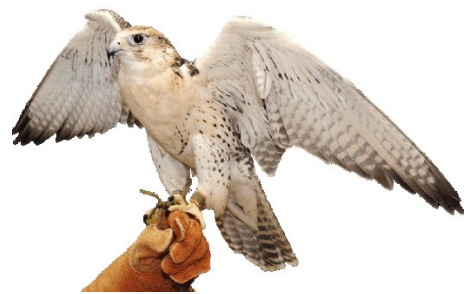

Fig. 1. UAE Falcon

### B. Falcon Diseases

Falcon diseases encompass a range of health issues that can impact the well-being of these birds of prey. One common ailment is 'Liver' disease, often attributed to nutritional imbalances or infectious agents. Another notable disease is 'Aspergillosis' a fungal infection caused by the Aspergillus species. Both diseases emphasize the importance of vigilant monitoring, early intervention, and effective management strategies in falconry practices.

*1. Liver Disease:* Falcon liver disease is a condition that affects the liver of falcons, and it can result from various factors such as nutritional imbalances or infectious agents. The disease may manifest with symptoms like hepatomegaly (enlarged liver), jaundice, and altered behavior in falcons. The impact of liver disease on falcons can be severe, affecting their overall health and performance. As the liver plays a crucial role in metabolism, detoxification, and digestion, any impairment in its function can lead to a cascade of health issues. Falcons with liver disease may exhibit reduced energy levels, compromised immune function, and impaired hunting or flying abilities. Prompt diagnosis and appropriate treatment are essential to mitigate the impact of liver disease.

*2. Aspergillosis Disease:* Falcon Aspergillosis is a respiratory disease caused by the inhalation of Aspergillus spores, typically by falcons. Aspergillus is a common fungus found in the environment, and when inhaled, it can lead to the development of respiratory infections. Falcons affected by Aspergillosis may exhibit symptoms such as difficulty breathing, lethargy, decreased appetite, and nasal discharge. The impact of Aspergillosis on falcons can be severe, potentially leading to chronic respiratory problems and compromised flight performance. In advanced cases, the disease can be fatal. The fungus often targets the respiratory tract, causing inflammation and compromising the bird's overall respiratory function. Early detection and prompt treatment are crucial for managing Aspergillosis in falcons. Veterinarians and falconers play a pivotal role in monitoring bird health, implementing preventive measures, and providing appropriate medical interventions to mitigate the impact of this respiratory disease on the affected birds.

*C. Deep Learning Model*

Deep learning, a subset of machine learning, entails training artificial neural networks to execute tasks without explicit programming. It is defined by the employment of deep neural networks, comprising multiple layers of interconnected nodes that enable the model to autonomously acquire hierarchical representations from data. These models have shown notable success in classifying diseases across diverse domains, presenting substantial progress in medical diagnostics. Their efficacy in disease identification and classification arises from their capacity to automatically learn hierarchical representations from data. In the realm of bird disease classification, deep learning models are transforming approaches by providing swifter, more precise, and less invasive methods when juxtaposed with traditional techniques.

*1. ConvNeXt Model:* ConvNeXt, a deep learning model developed by Google Research, aims to improve feature learning using mixed depthwise separable convolutions. It achieves a balance between network depth and parameters by introducing cardinality, which involves splitting channels into groups to enhance connectivity efficiency. Connectivity patterns like grouped convolution and global average pooling contribute to effective feature capture. The model's versatility is evident in tasks like image classification and object detection. However, the architectural complexity of ConvNeXt may result in increased computational requirements, necessitating careful parameter tuning. Despite these considerations, ConvNeXt's innovative features position it as a robust choice for tasks that demand nuanced feature representations in deep learning applications.

*2. EfficientNet Model:* EfficientNet, an innovative convolutional neural network architecture devised by Mingxing Tan and Quoc V. Le, transforms the scaling of models to enhance efficiency in image classification. This architecture employs a compound scaling strategy that uniformly adjusts depth, width, and resolution, achieving an optimal balance between computational complexity and model accuracy. EfficientNet utilizes efficient blocks containing inverted residuals and MBConv blocks, introducing parameters for depth, width, and resolution. The Swish activation function replaces the conventional ReLU, contributing to enhanced training and generalization. Importantly, EfficientNet attains state-of-the-art performance while utilizing fewer parameters, establishing it as a powerful choice for diverse computer vision tasks and environments with resource constraints, where the simultaneous optimization of efficiency and accuracy is crucial.

The paper is organized as follows: Section 2 offers an extensive literature review concentrating on systems for classifying falcon diseases. Section 3 details the methodology employed to assess the proposed concatenated deep learning model. Section 4 presents the results and conducts a performance analysis. Lastly, Section 5 serves as the conclusion, summarizing the research contributions articulated in the paper.

II. LITERATURE REVIEW

The study presented in [1] focuses on ingluvitis, aspergillosis, and bacterial enteritis as predominant health issues affecting falcons in the UAE. Emphasizing the potential transmission of these infections to falcon owners, the importance of regular health check-ups and disease control measures is underscored. In [2], a novel deep learning approach is proposed for automatic falcon disease recognition, utilizing multi-scale feature fusion to enhance accuracy and robustness in diagnosing various falcon diseases from radiographs. [3] investigates the effectiveness of pre-trained convolutional neural networks in classifying bird diseases from images, assessing architectures such as VGG16, ResNet50, and DenseNet169 for identifying diseases like Newcastle disease, Marek's disease, and avian influenza. In [4], a highly accurate and efficient method is developed for automatic bird species recognition from images, incorporating a hybrid deep learning architecture with a pre-trained VGG16 network for feature extraction, an optional Long Short-Term Memory network for capturing temporal dependencies in bird vocalizations, and a softmax classifier for predicting bird species. Finally, [5] discusses an effective method for bird species classification using deep convolutional neural networks and transfer learning, focusing on the utilization of pre-trained CNNs such as ResNet and VGG16. In summary, the literature consistently supports the effectiveness of concatenated deep learning models for falcon diseases classification.

## III. PROPOSED METHODOLOGY

The proposed concatenated deep learning model, combining the ConvNeXt and EfficientNet Models, offers advantages for disease classification applications by leveraging the distinctive strengths of both architectures. This concatenated model aims to improve feature extraction and classification capabilities, capitalizing on ConvNeXt's integration of depthwise and pointwise convolutions that enhance channel connectivity, complementing EfficientNet's efficiency in scaling network dimensions. This synergistic approach offers a comprehensive method to capture a wider range of features and patterns within medical data, ultimately leading to enhanced accuracy in disease classification. The methodology for classifying various falcon diseases employs this advanced AI model, commencing with the curation of a comprehensive dataset of endoscopy images representing conditions like 'Normal,' 'Liver' disease, and 'Aspergillosis.' Transfer learning techniques facilitate efficient model training, and the hybrid architecture is fine-tuned on the dataset to optimize its ability to discern subtle patterns indicative of specific diseases.

The pseudocode outlining the proposed concatenated AI model approach for falcon diseases classification is presented below:

- begin

– Split the labeled falcon endoscopy dataset into two groups: training (80%) and testing (20%).

– Construct the concatenated deep learning model incorporating features from ConvNeXt and EfficientNet.

– Define the input (first) layer of the concatenated AI model as 128 x 128.

– Specify the output (last) layer of the concatenated AI model to include three classes [Normal, Liver, Aspergillosis].

– Train the concatenated AI model using the training dataset and validate its performance.

– Utilize the test dataset to predict falcon endoscopy types and assess performance metrics such as accuracy, precision, recall, and F1-score.

- end

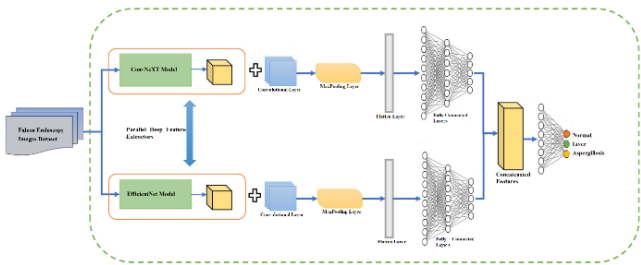

Fig. 2. Proposed Concatenated AI Model

## IV. RESULTS AND ANALYSIS

In this study, we conducted a thorough evaluation of proposed concatenated models through extensive experiments. The assessment utilized falcon endoscopy images sourced from the Sharjah Falcon Clinic, comprising a dataset of 610 labeled images representing three distinct falcon disease classes. The distribution of falcon diseases is illustrated in Figure 7x. To establish robust training, testing, and validation sets, we randomly divided this dataset. The training process, conducted on the Python 3 Google Compute Engine backend of Google Colab Pro, featured GPU support and 12.7 GB System RAM. With 80% of the dataset allocated for training and the remaining 20% for internal validation, the training phase spanned 50 epochs, each with a batch size of five. Following an intensive training regimen, model performance underwent rigorous evaluation on the test set, while hyperparameter fine-tuning was executed using the validation set.

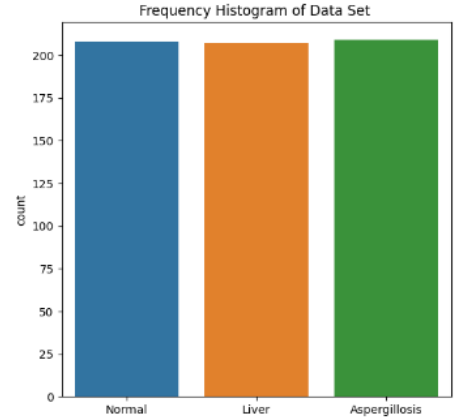

Fig. 3. Falcon Dataset distribution

To evaluate the model's performance, metrics such as accuracy, precision, recall, f1-score and ROC curve are employed, providing a comprehensive assessment of its diagnostic capabilities. The experimental results demonstrate the high accuracy and reliability of the proposed concatenated AI model solution in classifying various falcon diseases. The implemented falcon diseases classification system, utilizing the concatenated AI model, has undergone performance evaluation. The model demonstrates a notable training accuracy of 99.65%, with a corresponding validation accuracy of 98.50%. In terms of losses, the model achieves a training loss of 0.0865 and a validation loss of 0.0773.

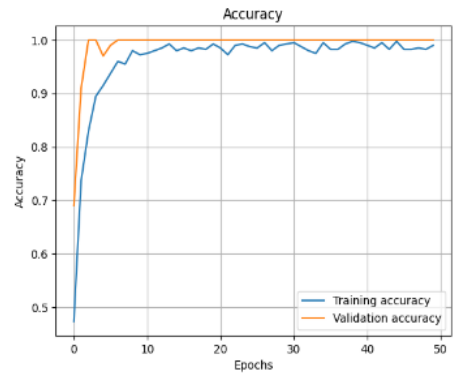

Fig. 4. Accuracy plots of the concatenated AI models

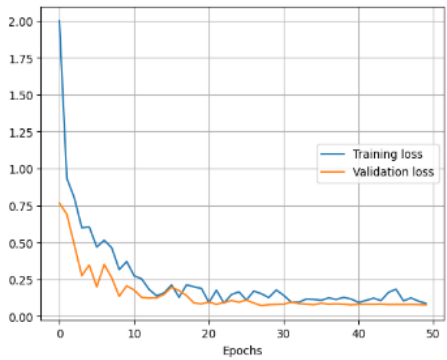

Fig. 5. Loss plots of the concatenated AI models

To evaluate the model's performance, metrics such as accuracy, precision, recall, f1-score and ROC curve are employed, providing a comprehensive assessment of its diagnostic capabilities. The experimental results demonstrate the high accuracy and reliability of the proposed concatenated AI model solution in classifying various falcon diseases. The implemented falcon diseases classification system, utilizing the concatenated AI model, has undergone performance evaluation. The model demonstrates a notable training accuracy of 99.65%, with a corresponding validation accuracy of 98.50%. In terms of losses, the model achieves a training loss of 0.0865 and a validation loss of 0.0773.

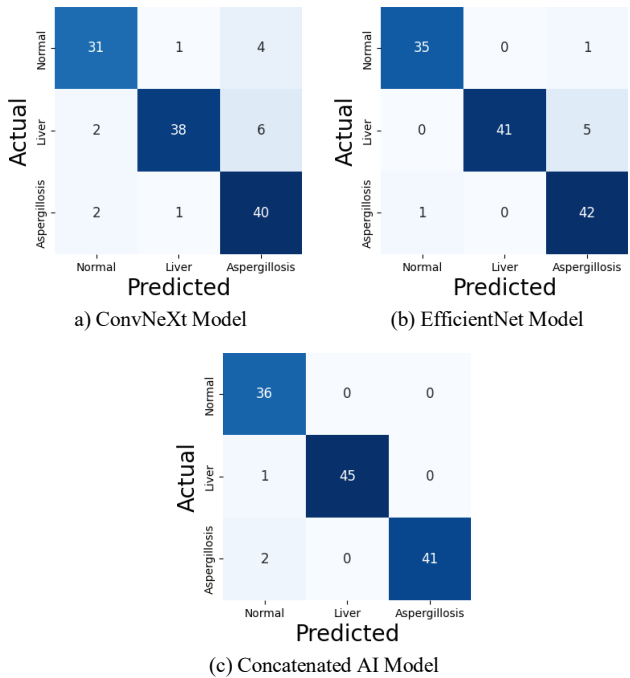

a) ConvNeXt Model
(b) EfficientNet Model
(c) Concatenated AI Model

Fig. 6 AI Models Confusion Matrix

The confusion matrix proves to be a potent and flexible tool in deep learning-based classification tasks, providing a comprehensive breakdown of model performance that extends beyond basic accuracy metrics. Its capacity to pinpoint specific areas for improvement and inform decision-making establishes it as a crucial element in evaluating and refining deep learning models. Presented in Fig. 6, the confusion matrix illustrates the ConvNeXt Model, EfficientNet Model, and Concatenated ConvNeXt-EfficientNet Model's performance based on 126 test images across diverse classes. Comparative analysis against state-of-the-art models (Table I) underscores the superior performance of the proposed concatenated AI model. Notably, this model demonstrates heightened prediction accuracy compared to other models examined in the study. The experimental results emphasize the potential of the proposed concatenated deep learning techniques in classifying falcon diseases, underscoring the ongoing need for research and refinement to further enhance the model's reliability and efficiency. Fig. 7 displays some examples of outcomes generated by the network. Each image is accompanied by the network's forecast, and the accuracy of the prediction is determined by its correspondence to the label within the parentheses.

TABLE I. PERFORMANCE EVALUATION OF AI CLASSIFIERS

| AI Models | Types | Accuracy | Precision | Recall | F1-Score |
|---|---|---|---|---|---|
| ConvNeXt | Normal | 0.87 | 0.89 | 0.86 | 0.87 |
| | Liver | 0.89 | 0.95 | 0.83 | 0.88 |
| | Aspergillosis | 0.86 | 0.80 | 0.93 | 0.86 |
| | Average | 0.87 | 0.88 | 0.87 | 0.87 |
| EfficientNet | Normal | 0.97 | 0.97 | 0.97 | 0.97 |
| | Liver | 0.95 | 1.00 | 0.89 | 0.94 |
| | Aspergillosis | 0.93 | 0.88 | 0.98 | 0.92 |
| | Average | 0.94 | 0.95 | 0.95 | 0.94 |
| Proposed Concatenated Model | Normal | 0.96 | 0.92 | 1.00 | 0.96 |
| | Liver | 0.99 | 1.00 | 0.98 | 0.99 |
| | Aspergillosis | 0.98 | 1.00 | 0.95 | 0.98 |
| | Average | 0.98 | 0.97 | 0.98 | 0.98 |

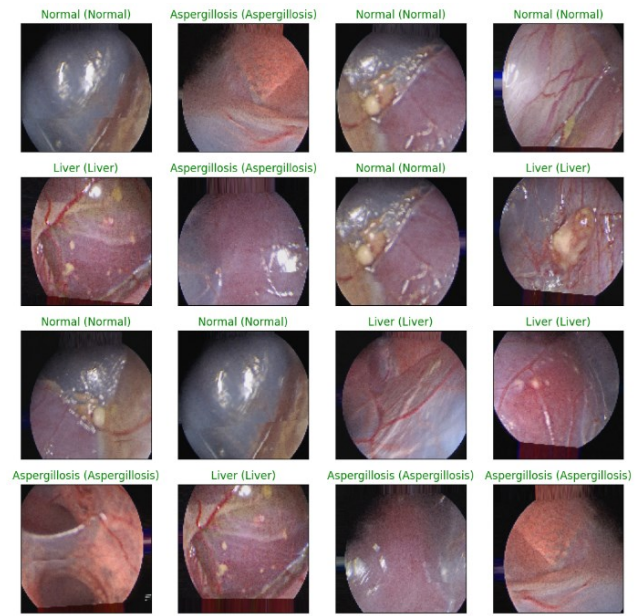

Fig. 7. Prediction performance of the Concatenated AI model

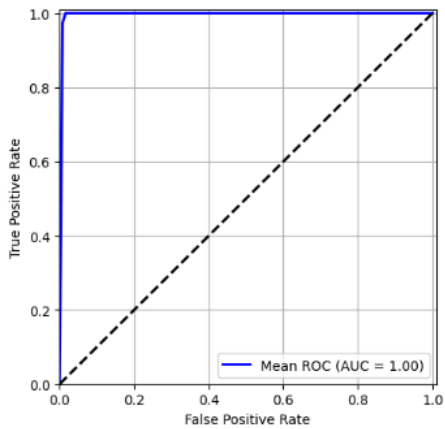

Fig. 8. ROC curve of the proposed concatenated AI model

ROC curve is significant in deep learning-based classification tasks as it provides a nuanced and visually interpretable performance assessment, aiding in the optimization of decision thresholds and facilitating better understanding of a model's discriminatory power across different scenarios. The area under the ROC curve serves as a quantitative measure of the model's overall discriminatory power, with higher AUC values indicating better performance. Fig 8 show the ROC curve of the proposed concatenated AI model.

## V. Conclusion

This paper introduces an advanced concatenated AI deep learning model specifically designed for classifying falcon diseases, showcasing superior accuracy compared to existing methods. Leveraging transfer learning enhances its feature extraction and classification capabilities. The proposed model is comprehensively presented and evaluated for falcon diseases classification, with a comparative analysis against ConvNeXt and EfficientNet models. Evaluation metrics encompass accuracy, precision, recall, F1-score, and a confusion matrix. Remarkably, the concatenated AI model achieves testing accuracies of 98%, surpassing other deep learning models considered in this study. Future efforts will focus on developing lightweight models to further improve performance and extending the model's application to broader disease analysis tasks.